# A Composable Multimodal Framework for cine CMR-Text–Driven Prediction of Heart Failure Outcomes

**Jianzhou Chen[1,§], Jinyang Sun[2,§], Xiumei Wang[3], Xi Chen[4], Heyu Chu[5], Guo Song[5], Yuji Luo[5], Xingping Zhou[6,*], and Rong Gu[1,*]**

[1]Department of Cardiology, Nanjing Drum Tower Hospital, State Key Laboratory of Pharmaceutical Biotechnology, Nanjing University, Nanjing 210008, China
[2]School of Electronic Information and Electrical Engineering, Shanghai Jiao Tong University, Shanghai 200240, China
[3]College of Electronic and Optical Engineering, Nanjing University of Posts and Telecommunications, Nanjing 210003, China
[4]College of Integrated Circuit Science and Engineering, Nanjing University of Posts and Telecommunications, Nanjing 210003, China
[5]Department of Cardiology, Nanjing Drum Tower Hospital Clinical College of Nanjing Medical University, Nanjing 210008, China
[6]Institute of Quantum Information and Technology, Nanjing University of Posts and Telecommunications, Nanjing 210003, China

§ These authors contributed equally to this work.
E-mail: *zxp@njupt.edu.cn* and *gurong.nju@163.com*

## Abstract

*Objective.* Heart failure is one of the leading causes of death worldwide, with millions of deaths each year, according to data from the World Health Organization (WHO) and other public health agencies. While significant progress has been made in the field of heart failure, leading to improved survival rates and improvement of ejection fraction, there remains substantial unmet needs, due to the complexity and multifactorial characteristics. This study aims to propose and evaluate a composable strategy framework for assessment and treatment optimization in heart failure, designed to provide more holistic patient evaluation and management. *Approach.* The framework leverages multi-modal algorithms to analyze a comprehensive range of patient data, explicitly integrating cine cardiac magnetic resonance (cine CMR) sequences, structured clinical metrics (e.g., lab results, demographics), and unstructured textual records (e.g., medical history, prescriptions). By integrating these various data sources, our framework offers a more holistic evaluation and optimized treatment plan for patients. *Main results.* The multi-modal framework demonstrates superior accuracy in HF prognosis prediction compared to single-modal AI algorithms. Additionally, it enables a detailed evaluation of the impact of various pathological indicators on HF outcomes. *Significance.* By integrating heterogeneous clinical data in a systematic manner, this approach supports more comprehensive prognosis assessment and facilitates optimized, personalized treatment planning for heart failure patients.

Keywords: Heart Failure, Multimodal Framework, Prognosis Prediction, Dynamic Attention Mechanism, Personalized Treatment

---

## 1. Introduction

Heart failure (HF) is a complex and multifactorial disease, and its clinical manifestations, treatment and prevention are widely challenging. With the aging of the population and the increase in cardiovascular disease, the incidence rate of heart failure continues to rise, resulting in a huge consumption of medical resources worldwide. In the treatment of heart failure, lifestyle intervention, pharmacological management, the device-based management, and heart transplantation are the main treatment methods. The addition of a neprilysin inhibitor to renin–angiotensin system inhibition (ARNI), and the advent of SGLT2 inhibition are the main supplemental solutions of pharmacological management (Lam *et al* 2023). The incidence rate of newly diagnosed heart failure has reportedly stabilized or may even be declining in the general population. People with heart failure also live longer. However, the long-term trend of increasing numbers of individuals at risk for heart failure and increasing survival rates for individuals with heart failure has led to an overall increase in the prevalence of heart failure patients.

On the other hand, the vigorous development of deep learning has brought new paradigms to scientific research in various fields, and it has also been widely used in clinical studies (Rajpurkar *et al* 2022). Deep learning can discover patterns, predict diseases, assist in diagnosis, and optimize treatment plans by analyzing large amounts of medical data, significantly improving medical efficiency. It has played a significant role in medical image analysis (Shen *et al* 2017), clinical decision support (Tayebi Arasteh *et al* 2024), clinical data analysis (Chen *et al* 2022, Mekkes *et al* 2024), and other fields.

Recent years, scientists employ machine learning to mine the mapping relationship between HF and various important factors. For instance, deep learning is utilized to identify high-risk patients based on specific comorbidities such as subclinical hypothyroidism (Yagi *et al* 2023) and genetic markers like clonal hematopoiesis (Chen *et al* 2023). Additionally, approaches utilizing general risk assessment (Adler *et al* 2020, Kokori *et al* 2024), distinct patient phenotypes (Segar *et al* 2024, Monzo *et al* 2024), structural analysis (Lau *et al* 2023), and clinical biomarkers (Lumish *et al* 2024, Ma *et al* 2022) are widely used to predict composite outcomes like hospitalization and Cardiovascular death (CV death) (Flores *et al* 2021). However, most of these works focus on single factors, which may lead to partial judgment. Fortunately, multimodal artificial intelligence (AI) has emerged to process diverse data types simultaneously (Yin *et al* 2024). This technology is rapidly applied in clinical studies, particularly in neurology for Alzheimer's assessment (Qiu *et al* 2022, Zhou *et al* 2025), brain injury prognosis (Rohaut *et al* 2024), and dementia diagnosis (Xue *et al* 2024). Multimodal AI demonstrates superior performance in oncology and pathology, covering areas such as radiation oncology (Oh *et al* 2024), clinical study automation (Tayebi Arasteh *et al* 2024), recurrence prediction (Volinsky-Fremond *et al* 2024), and advanced pathological analysis (Lu *et al* 2024, Klughammer *et al* 2024). Moreover, it has extended into chronic disease management, such as diabetes care (Li *et al* 2024). These studies reveal that integrating images, text, and numerical data provides more accurate clinical reasoning than single-modal systems.

To address these challenges, we propose the Multimodal Post-Recovery Tracking Model (M-PRTM), a composable strategy framework designed to systematically integrate cine CMR sequences, structured clinical metrics, and textual medical records for HF prognosis prediction. M-PRTM combines structured data (such as patient demographics and clinical metrics) and unstructured textual records (including medication history) with cine CMR sequences to create a comprehensive model. Our framework employs specialized models for each data modality, including the Dual Attention-guided Efficient Transformer model (DAE-Former) (Azad *et al* 2023) for cine CMR data, a fully connected network for structured data, and a BERT-based encoder for textual data. These features are then fused using a Self-Reasoning Attention mechanism, which dynamically prioritizes the most important information for predictions, such as drug prescriptions and vital signs. Our results show that M-PRTM achieves 96.5% accuracy in predicting clinical outcomes, outperforming traditional single-modal approaches. The framework's ability to dynamically re-weight modalities via the composable attention strategy enables more reliable predictions, while its multimodal nature supports personalized treatment strategies and early intervention. Through rigorous testing with data from 136 imaging cases and 688 clinical records, M-PRTM demonstrates its effectiveness in predicting HF prognosis, offering a scalable solution for optimizing patient care and improving outcomes.

## 2. Methods

### 2.1 M-PRTM's Overview

We have described the underlying structure that supports the data modalities in Fig. 1. In the Data column, patient data are categorized into three main types: numerical indicators, textual prescriptions, and cine CMR sequences. Numerical indicators include demographic information (e.g., gender and age) and key clinical metrics (e.g., blood pressure and BMI). Textual records primarily consist of prescription information, such as drug type, dosage, and treatment stage. Meanwhile, cine CMR data capture anatomical and functional features of the myocardium, including regions of fibrosis. In particular, Late Gadolinium Enhancement (LGE) is a key imaging biomarker for detecting myocardial fibrosis. It provides crucial insights into the structural integrity and scarring of the myocardium, which are closely linked to the progression of HF (Gonzalez and Kramer 2015, López *et al* 2021). To leverage these imaging insights, we extract vector features from numerical, textual, and cine CMR for multimodal fusion-based predictions. Specifically, we provide an example in the Medical Prescription section. Considering that drug types and dosages often vary across time or treatment stages during clinical practice, our data processing accounts for these dynamic changes to enhance the M-PRTM's adaptability to patient medication patterns. The comprehensive workflow for M-PRTM's development and validation is presented in Fig. 1.

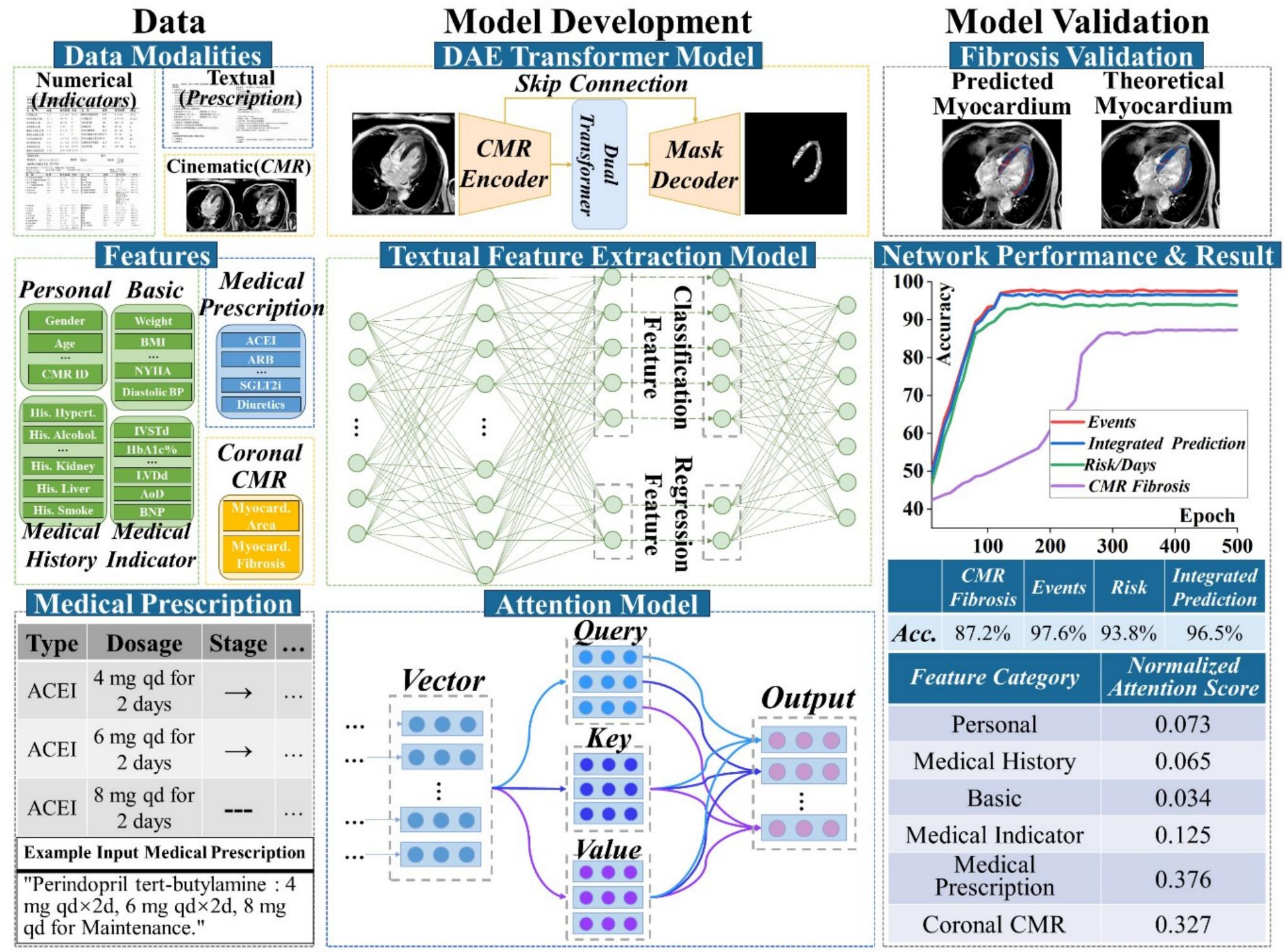

**Figure 1.** A M-PRTM's development and validation, including overview of data, model development, and model validation.

The Model Development column outlines three modules designed to process imaging, numerical, and textual tasks. For imaging data, we employ DAE-Former (Azad *et al* 2023). The model integrates a skip-connected CMR encoder and masked decoder, which enable precise predictions of myocardial fibrosis regions and can significantly improve segmentation accuracy. For the numerical type data, we construct a feature extraction model to map the numerical inputs into classification and regression features. Those features are further refined using an Attention Model, which calculates the "Query-Key-Value" weights to highlight critical attributes impacting prediction outcomes. For the textual type data, we apply a pre-trained Bidirectional Encoder Representations from Transformers (BERT) model (Devlin *et al* 2019, Koroteev 2021) to embed textual information. It is complemented by an Attention Model to dynamically adjust the importance of textual features, effectively capturing the impact of phased drug adjustments on prediction outcomes.

In the Model Validation step, the Fibrosis Validation module evaluates the predicted fibrosis regions, showing high consistency between the prediction results and the theoretical annotations. This demonstrates the accuracy and efficacy of the model in fibrosis segmentation. Additionally, the Network Performance and Result curves illustrate the model's accuracy across multiple prediction tasks, including fibrosis segmentation, event prediction, risk prediction, and integrated prediction, achieving 87.2%, 97.6%, 93.8%, and 96.5%, respectively. Finally, modality contribution analysis reveals that therapeutic agents and cine CMR contribute the most to the predictions, with importance scores of 0.376 and 0.327, respectively. Other feature categories, such as numerical indicators and medical indicator, also make contributions but with lower importance scores.

Overall, the validation results can confirm that our multimodal framework effectively integrates CMR, numerical type, and textual type data, enabling highly accurate and reliable predictions of clinical outcomes. This highlights the potential of the M-PRTM to support personalized medicine and optimize patient management strategies.

## 2.2 Data Collection and Processing

### 2.2.1 Data Collection

Our research focuses on HF prognosis prediction and myocardial fibrosis detection. We retrospectively assemble cine CMR data from 136 patients and structured clinical records from 688 patients at Nanjing Drum Tower Hospital. The clinical dataset includes demographics, vital signs, comorbidities, medications, laboratory indices, and CMR-derived measurements, as summarized in Table 1.

**Table 1.** Baseline Characteristics and Key Clinical Indicators

| Column Name | Mode Type | Mean | Standard Deviation | Range |
|---|---|---|---|---|
| Gender | Numerical | 0.8014 | 0.4003 | (0,1) |
| Age | Numerical | 52.367 | 14.391 | (18,79) |
| Weight | Numerical | 73.006 | 16.877 | (42.0,121.2) |
| Heart Rate (bpm) | Numerical | 85.602 | 19.331 | (43,153) |
| … | … | … | … | … |
| Diastolic BP | Numerical | 85.242 | 18.890 | (39,144) |
| Myocardial Infarction | Numerical | 0.1176 | 0.3233 | (0,1) |
| Pacemaker | Numerical | 0.0441 | 0.2061 | (0,1) |
| Medical Indicator | Text | — | — | — |
| IVSTd | Numerical | 0.9391 | 0.2253 | (0.5,1.87) |
| ALT | Numerical | 38.265 | 42.408 | (5.3,340.8) |
| AST | Numerical | 30.633 | 26.288 | (10.8,235) |
| Albumin/Globulin Ratio | Numerical | 1.6280 | 0.2911 | (0.76,2.32) |
| … | … | … | … | … |
| LDL-C | Numerical | 2.5078 | 0.9104 | (0.46,5.07) |
| Apo-A1 | Numerical | 0.9065 | 0.2043 | (0.46,1.63) |
| HbA1c% | Numerical | 6.2455 | 1.1636 | (4.3,11.3) |

*Note:* "Medical Indicator" represents specific textual diagnostic labels extracted from clinical notes (e.g., 'Hypertension', 'Diabetes Type 2').

Inclusion criteria require adults over 18 years with available cine CMR studies or complete clinical records within the study window. Exclusion criteria remove duplicate entries, missing primary outcomes, and CMR studies with non-diagnostic image quality. Where available, imaging and clinical records link through anonymized identifiers to enable multimodal analyses while protecting patient privacy. We present an example of the data organization and outcome recording in Table 2.

**Table 2.** Example of structured clinical records and outcome tracking

| Patient Info | | | | Cardiac Death | | Rehospitalization | | MACCE | | |
|---|---|---|---|---|---|---|---|---|---|---|
| Gender | Age | Discharge Date | Follow-up (Days) | Event | Days | Event | Days | Event | Cause | Days |
| Female | 33 | 2021/01/25 | 810 | 0 | - | 1 | 344 | 1 | Worsening HF | 341 |
| Male | 68 | 2020/05/22 | 1087 | 1 | 1087 | 1 | 1087 | 1 | Fatal MI | 1087 |
| Male | 30 | 2019/04/22 | 1447 | 0 | - | 0 | - | 0 | - | - |

The overall dataset (*N*=688) contains 93 cases of cardiac death (13.52%), 158 cases of rehospitalization (22.97%), and 143 MACCE events (20.78%). The data is split into Training (70%), Validation (10%), and Test (20%) sets. Specifically, the Training Set (*n*=482) includes 65 cardiac deaths, 111 rehospitalizations, and 100 MACCE; the Validation Set (*n*=69) contains 9 cardiac deaths, 16 rehospitalizations, and 14 MACCE; and the Test Set (*n*=137) comprises 19 cardiac deaths, 31 rehospitalizations, and 29 MACCE. For the fibrosis detection task, we use available manual annotations as reference masks and evaluate predicted regions against these annotations. Regarding the Clinical Endpoints and Prediction Tasks, the primary clinical endpoint is the composite outcome of Major Adverse Cardiovascular and Cerebrovascular Events (MACCE), defined as the occurrence of heart failure rehospitalization, cardiovascular death, or non-fatal stroke within a 24-month prediction window. Accordingly, the prediction framework operates in a cascaded manner: first, a binary classification task predicts the probability of MACCE occurrence wwithin the 24-month prediction window; subsequently, for patients predicted as high risk, an auxiliary regression branch estimates the time-to-event in days. The regression target is defined as the interval between discharge and the first observed MACCE event. Regression performance is evaluated on the event-positive subset with valid event-time annotations. This regression branch is intended to provide a supplementary estimate of event timing for high-risk patients rather than to replace formal survival modeling In terms of cohort statistics, the study includes a total of 688 patients. The follow-up duration covers a substantial time span, with a median of 895 days and a mean of 931 days. Regarding the event distribution, the overall dataset contains 143 MACCE events (20.78%), representing the positive cases, versus 545 negative cases. This distribution informs our choice of evaluation metrics, specifically the inclusion of the area under the precision–recall curve (AUPRC) to account for class imbalance.

To prevent data leakage and ensure a realistic prospective prediction scenario, we enforce a strict temporal protocol where the prediction baseline is defined as the date of discharge. Input features are rigorously censored to include only data available at this time point: textual records are limited to those generated prior to or on the discharge date, and cine CMR data is restricted to the baseline scan from the index hospitalization. Any post-discharge longitudinal data is explicitly excluded to ensure the model predicts the 24-month MACCE outcomes using only prospectively available information. Following these definitions, data preprocessing is performed where we standardize cine CMR volumes to 512 × 512 × 16 and tokenize free text with a BERT tokenizer, then pad or truncate to 512 tokens with attention masks. In addition, we impute missing numerical values with the median and apply min–max scaling, and we mitigate class imbalance with under-sampling. Ethical approval is granted by the Ethics Committee of Nanjing Drum Tower Hospital (Approval No. 2025-0094-015), and consent to participate is not applicable.

#### 2.2.2 Cine CMR Processing

All cine CMR are standardized in spatial resolution and uniformly scaled to a size of 512×512×16 pixels. To mitigate intensity variations, the intensity values are normalized to ensure consistent preprocessing standards across all samples.

#### 2.2.3 Textual Data Processing

A Transformer-based tokenizer is used for encoding. The BERT tokenizer tokenizes textual data, truncating or padding all text to a fixed length of 512 tokens for consistency. Additionally, an attention mask is generated to handle texts of varying lengths. The encoded textual features are then converted into tensors for the input.

#### 2.2.4 Numerical Data Processing

The median is used to impute missing values, minimizing the influence of extreme values. Subsequently, all numerical features are scaled to the range from 0 to 1, using Min-Max normalization, ensuring equal weighting of different features during training. To address class imbalance in minority case data, an under-sampling strategy is applied. This approach enhances the model's focus on minority class samples, resulting in a more balanced data distribution and improved predictive performance for minority categories while maintaining overall stability.

### 2.3 M-PRTM Architecture

Our research establishes a multimodal feature extraction and fusion framework. The proposed method integrates the characteristics of cine CMR, textual, and numerical data, introducing the targeted data processing workflows and the model structures. The methodology includes three main stages: data collection, single-modal feature extraction (cine CMR features, numerical features, and textual features), and multimodal feature fusion, as illustrated in Fig. 2.

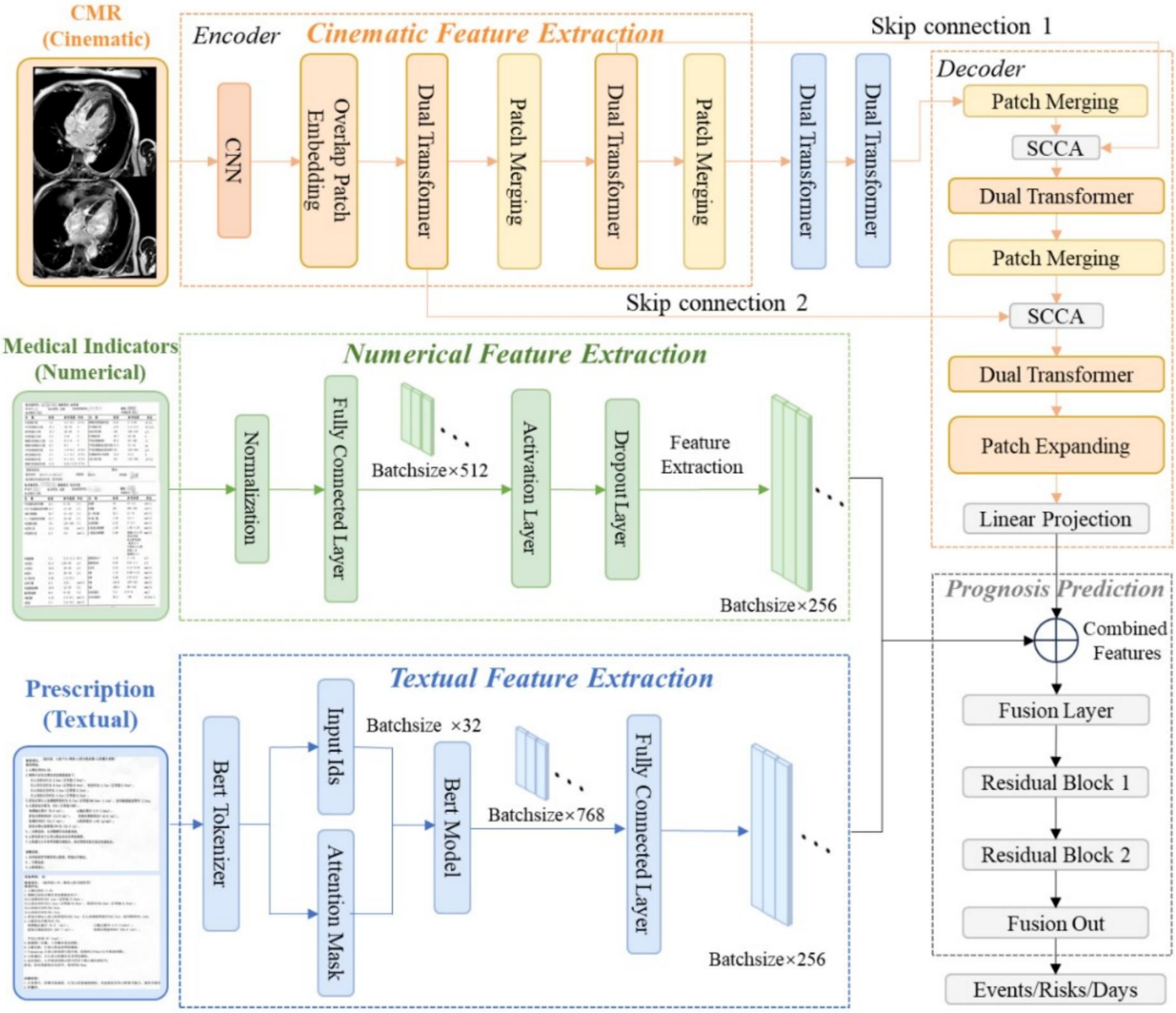

**Figure 2.** Framework of Multimodal Feature Extraction and Fusion. The diagram illustrates the feature extraction process for cine CMR, numerical, and textual data, followed by multimodal feature fusion. Cine CMR features are extracted using the DAE-Former with dual attention and skip connections, numerical features are processed through fully connected layers, and textual features are encoded using BERT. The extracted features are fused through a fusion layer and residual blocks to predict prognosis outcomes such as events, risks, and follow-up days.

### 2.3.1 Cine CMR Feature Extraction Module

In myocardial fibrosis detection tasks, we employ the DAE-Former model (Azad *et al* 2023). DAE is a pure Transformer architecture designed specifically for medical image segmentation. It addresses the limitations of traditional convolutional neural networks (CNNs) in capturing long-range dependencies and global contextual information while overcoming the high computational complexity of standard Transformers. The model consists of an input encoder, output encoder, Efficient Dual Attention (Efficient Attention and Transpose Attention), Skip Connection Cross Attention (SCCA), and an output layer. The core components of Efficient Dual Attention and SCCA are detailed in Fig.3 with their structural diagrams. In the structural diagrams of Figure 3, the symbol $\otimes$ denotes matrix multiplication, $\oplus$ denotes element-wise addition (residual connection), and $T$ represents the matrix transpose operation. *Norm* indicates Layer Normalization, and *FFN* refers to the Feed-Forward Network.

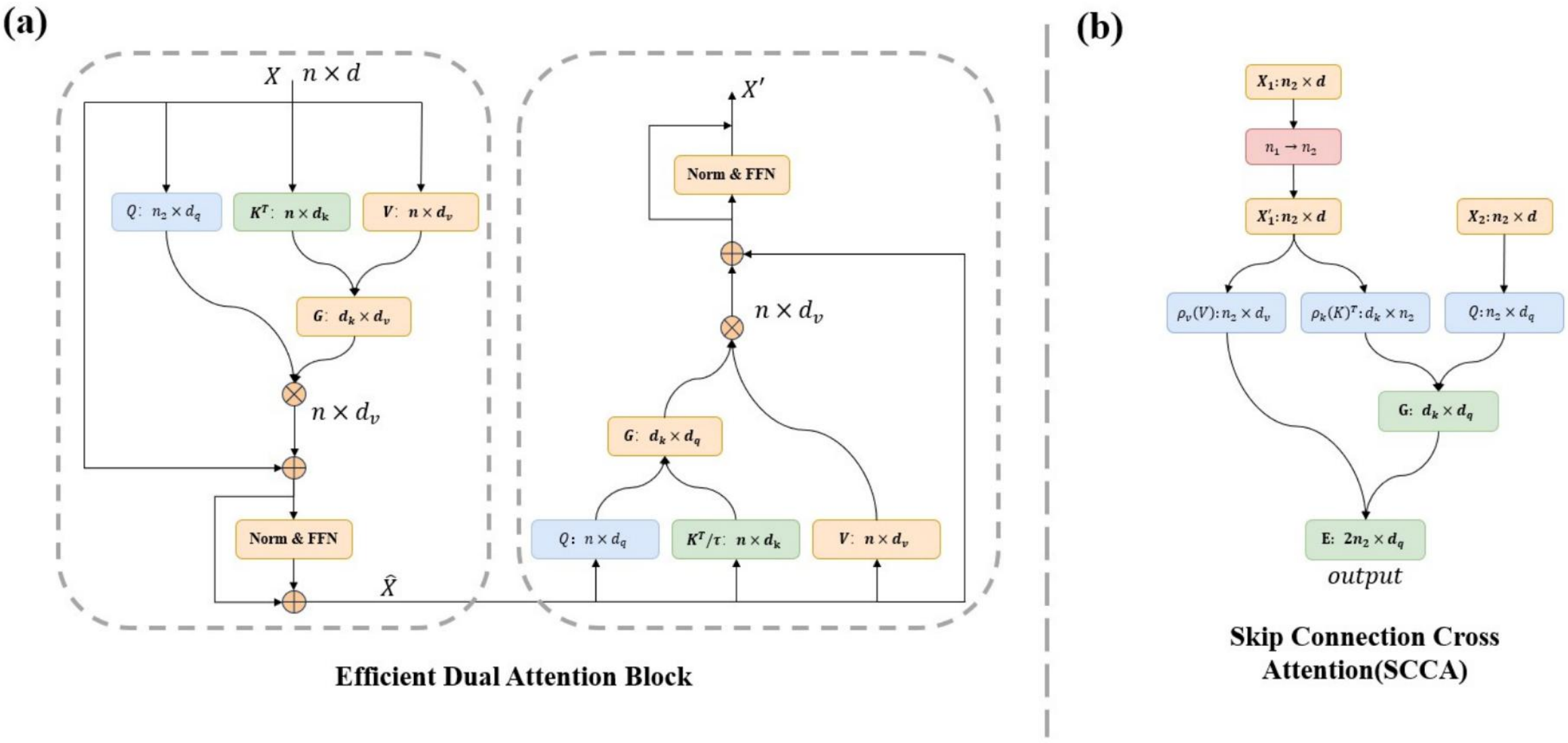

**Figure 3.** (a) Structure of the Efficient Dual Attention (EDA) module. (b) Structure of the Skip Connection Cross Attention (SCCA) module.

The EDA module is a dual Transformer block that integrates transposed (channel) attention and effective (spatial) attention, as shown in Fig. 3(a). This dual-attention block integrates effective attention, normalization, and transposed attention block that performs channel attention. Fig. 3(b) presents the structure of another key module SCCA, which employs a cross-processing approach to more effectively preserve low-level features. This module is capable of providing spatial information that helps recover fine-grained details when generating the output mask. Specifically, SCCA incorporates an effective attention mechanism. Its query input is derived from the encoder layer output forwarded through the skip connection, while the keys and values originate from the lower decoder layer. To integrate these two features, a linear layer is employed to scale to match the embedding dimension. The rationale for using as the query input is to model multi-level representations within the effective attention block.

### 2.3.2 Efficient Dual Attention

Efficient Dual Attention combines Efficient Attention (to capture spatial information) and Transpose Attention (to capture channel information), enhancing the model's ability to comprehend contextual features. These two mechanisms are applied sequentially, with normalization and residual connections incorporated to improve stability.

### 2.3.3 Skip Connection Cross Attention

Skip Connection Cross Attention (SCCA) introduces a cross-attention mechanism into the traditional U-Net architecture. It adaptively fuses encoder features with decoder low-level features, enabling seamless integration of encoder and decoder representations. This process involves linear projection and efficient attention fusion methods, which enhance the recovery of segmentation boundaries and fine details.

For myocardial fibrosis feature extraction from cine CMR images, the raw coronal CMR images are first analyzed frame by frame. Myocardial regions, fibrosis shadow regions, and other areas are segmented into different masks. The CMR images, matched with their corresponding masks, are then fed into the DAE-Former network. The M-PRTM identifies potential lesion areas within individual frames and tracks their consistency across multiple slices, leveraging the attention mechanisms inherent in Transformers. Consequently, our M-PRTM can generate stable and accurate predictions of lesion regions, with the ability to capture changes in lesion positions across consecutive frames.

### 2.3.4 Numerical Feature Extraction Module

In the numerical tasks, a fully connected layer is designed to process patients' clinical numerical indicators. The preprocessed numerical features are input into a fully connected network for deep feature extraction. The first layer maps the normalized input data into a 512-dimensional intermediate feature space, incorporating a ReLU activation function to introduce non-linearity. To prevent overfitting, a Dropout layer with a dropout rate of 0.2 is added after the first layer. The second layer further reduces the intermediate features to 256 dimensions to standardize the feature size, facilitating convenient input for subsequent multimodal feature fusion. The final output is a feature tensor with a size of batchsize × 256, consistent with features from other modalities.

### 2.3.5 Textual Feature Extraction Module

In this module, a pre-trained BERT model is employed to extract deep semantic features from the textual data. BERT can effectively capture the semantic relationships among words within a sentence, leveraging its strong contextual understanding capabilities. It constructs a comprehensive and accurate representation of textual features, which also enriches the predictive model with semantic insights. We employ BERT as a Pre-trained Language Model (PLM) to encode the semantic information from clinical text sequences.

As shown in Fig. 2, textual input is first processed through the BERT tokenizer. Patients' medical histories and medication records are integrated into natural language sentence forms and input into the BERT model for embedding. During this process, BERT utilizes its multi-layer bidirectional Transformer structure to parse the input text word by word, extracting dynamic semantic relationships for each word within its context. The model outputs a global feature vector that encapsulates the overall semantic information of the text, including key clinical details such as drug types and dosage adjustments.

To further enhance the expressiveness of textual features and align them with features from other modalities, a fully connected layer is added after the BERT model's output layer. This layer is designed to reduce dimensionality, mapping the 768-dimensional semantic vector to a 256-dimensional feature space. This dimensionality reduction not only decreases computational complexity but also ensures dimensional consistency with image and numerical features during the feature fusion stage. Besides, the step optimizes the representation of textual features, retaining original semantic information while improving computational efficiency.

### 2.3.6 Multimodal Feature Fusion

We adopt a multimodal fusion strategy incorporating an attention mechanism to integrate features from cine CMR, textual, and numerical modalities. To implement this, we design a fusion process that consists of two stages: independent modality feature extraction and multimodal feature fusion, as depicted in Fig. 2.

In the feature extraction stage, unique features are independently extracted from each modality through dedicated modules. cine CMR are processed by the DAE-Former model, capturing spatial and temporal information at the frame level. Textual data are processed by the BERT model, extracting semantic features to analyze medical histories and medication records. Numerical data are aggregated using a multi-layer perceptron (MLP) to represent clinical indicators. These modules ensure efficient learning of features within their respective spaces.

In the fusion stage, the model dynamically assigns weights to different modality features using an attention mechanism. Specifically, to formally define the Self-Reasoning Attention mechanism, let $F_i \in \mathbb{R}^d$ represent the input feature vector of the *i*-th modality (where $i \in \{Cinematic, Textual, Numerical\}$ ). The Query (*Q*), Key (*K*), and Value (*V*) matrices are computed through linear projections:

$$Q = FW^Q, \quad K = FW^K, \quad V = FW^V \tag{1}$$

where $W^Q, W^K, W^V \in \mathbb{R}^{d \times d_k}$ are learnable weight matrices. The attention weights are subsequently calculated using the scaled dot-product softmax function:

$$Attention(Q, K, V) = \text{softmax}\left(\frac{QK^T}{\sqrt{d_k}}\right)V \tag{2}$$

where $d_k$ is the scaling factor to prevent gradient vanishing. This mechanism dynamically assigns importance weights to different modalities based on their semantic relevance. These scores are scaled and transformed into probability distributions through the SoftMax function, which serves as weights for the feature fusion process. The weighted features are then combined into contextual features. This dynamic process evaluates the contribution of each modality, emphasizing features that are more critical for prediction tasks. We define the framework as 'Composable' due to the adaptive nature of the Self-Reasoning Attention module. Unlike static fusion models, this mechanism dynamically composes the final patient representation by re-weighting or masking specific modalities (Cine CMR, Textual, and Numerical) in real-time based on their availability and contribution to the prediction task. Additionally, the fusion strategy incorporates cross-modality regulation, enabling the interaction of information between modalities to optimize feature representation collaboratively. For example, cine CMR features can enhance the semantic representation of textual features, while textual features can improve the interpretation of numerical features. This dynamic interaction across modalities significantly boosts the expressiveness of fused features. Finally, the fused features are integrated through a linear layer to generate a unified feature representation for disease and mortality cause prediction. This multimodal fusion strategy not only leverages complementary information from different modalities but also achieves efficient predictive performance, providing essential support for personalized treatment strategies.

### 2.4 Implementation Details

To strictly prevent data leakage, dataset splitting is performed at the patient level, ensuring that all modalities (cine CMR, textual, and numerical) for a given patient reside exclusively in the same subset. We employ a stratified split strategy based on the distribution of MACCE events to maintain consistent class ratios across the Training (70%), Validation (10%), and Test (20%) sets. A fixed random seed (42) is used to ensure the reproducibility of the data splitting and training process. For model initialization, the textual branch utilizes pre-trained BERT weights to leverage robust semantic representations, while the cine CMR branch employs the DAE-Former architecture. The model training follows a deterministic approach to ensure consistent convergence and result replicability.

To mitigate overfitting given the limited dataset size, we adopt a Transfer Learning strategy by initializing encoders with pre-trained BERT and DAE-Former weights. Additionally, to address the modality mismatch (136 imaging vs. 688 clinical records), we implement a Modality Masking mechanism. Missing modality features are zero-padded and strictly ignored by the attention mask, enabling the model to be trained jointly on the entire cohort ($n$=688) rather than being restricted to the subset with complete imaging data. The model is implemented using the PyTorch framework. We utilize the AdamW optimizer with a learning rate of $1 \times 10^{-4}$ and a weight decay of $1 \times 10^{-2}$ to prevent overfitting. The batch size is set to 32, and the model is trained for 500 epochs. A cosine annealing scheduler is employed to dynamically adjust the learning rate during training.

Finally, the network is optimized using a composite loss function. The total loss $L_{total}$ is defined as the weighted sum of the prediction task loss (Cross-Entropy), the segmentation auxiliary loss (Dice Loss), and the auxiliary regression loss:

$$L_{total} = \lambda_{pred} L_{CE}(y, \hat{y}) + \lambda_{seg} L_{Dice}(M, \hat{M}) + \lambda_{reg} L_{MSE}(t, \hat{t}) \quad (3)$$

where $y$ and $\hat{y}$ denote the ground truth and predicted labels for the prognosis task, and $M$ and $\hat{M}$ represent the ground truth and predicted fibrosis masks, and $t$ and $\hat{t}$ denote the ground-truth and predicted event times for the auxiliary time-to-event regression task, respectively. The hyperparameters $\lambda_{pred}$, $\lambda_{seg}$ and $\lambda_{reg}$ balance the contribution of the three tasks during training. For the auxiliary regression branch, the prediction target is defined as the interval between discharge and the first observed MACCE event, and supervision is applied only to event-positive samples with valid event-time annotations. For the main classification metrics, 95% confidence intervals are estimated using bootstrap resampling on the independent test set.

## 3. Results

### 3.1 M-PRTM's Performance and Results

In this section, we present key performance metrics of the proposed M-PRTM alongside its prediction results. For the myocardial fibrosis detection task, the DAE-Former achieves stable training behavior and effective fibrosis feature extraction. Representative fibrosis prediction examples are shown in Fig. 4(a). The results indicate that the fibrosis regions predicted by DAE-Former closely align with manual annotations, achieving an accuracy of approximately 87%, demonstrating the model's effectiveness in processing imaging data. To validate the effectiveness of the DAE-Former in our specific myocardial fibrosis detection task, we compare its performance against mainstream medical image segmentation networks, including TransUNet and HiFormer. As presented in Table 3, our method achieves a Dice Similarity Coefficient (DSC) of 72.32% and a Hausdorff Distance (HD) of 21.66. Although the dataset size limits the absolute numerical performance, the DAE-Former outperforms the baseline methods, confirming its efficacy in accurately extracting fibrosis features for subsequent multimodal analysis.

**Table 3.** Comparison of Segmentation Performance with Baseline Methods

| Method | DSC(%)↑ | HD(pixels)↓ |
|---|---|---|
| TransUNet (Chen *et al* 2024) | 67.17 | 26.53 |
| HiFormer (Heidari *et al* 2023) | 69.66 | 18.58 |
| DAE-Former (Ours) | 72.32 | 21.66 |

A key contribution of our work is the design and implementation of the M-PRTM, which is specifically developed to integrate textual, numerical, and cine CMR for joint training. For the multimodal prognosis component, the M-PRTM also demonstrates reliable predictive performance. Representative examples of prognosis prediction are presented in Fig. 4(b), including mortality, causes of death, days to rehospitalization, and MACCE-related events. The results reveal strong alignment between predicted values and ground truth, highlighting the ability to handle incomplete recovery data and accurately predict patient outcomes. Furthermore, M-PRTM provides early recommendations for days to rehospitalization or follow-up visits, helping to prevent unnecessary mortality caused by deteriorating conditions during recovery. Fig. 4(c) highlights the unique functionality of M-PRTM in dynamically assessing patient risk and predicting recovery trajectories.

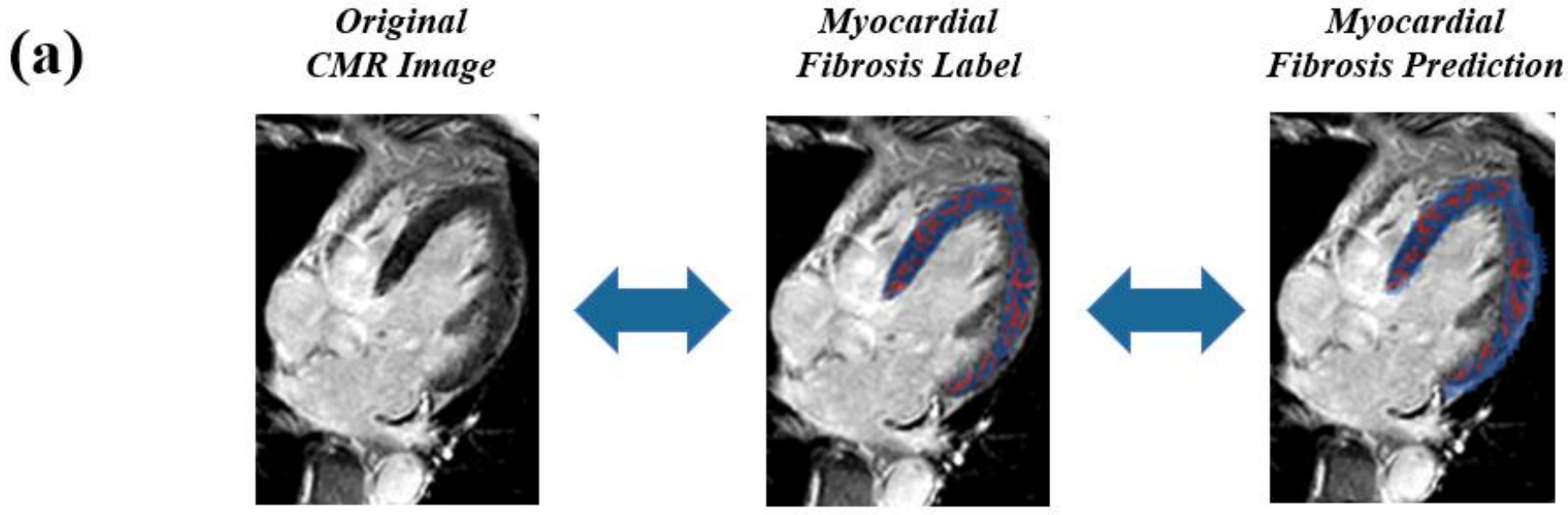

**(b)** **Example Of Prediction Results**

| | *Death (Days)* | *Cause Of Death* | *Cardiology Rehospitalization (Days)* | *MACCES Event (Days)* | *MACCES Cause* |
|---|---|---|---|---|---|
| *Ground Truth* | Yes (186) | Heart Failure | Yes (97) | Yes (Missing data) | Heart Failure Deterioration |
| *Predict* | Yes (173) | Heart Failure | Needed (104) | Yes (165) | Heart Failure Deterioration |

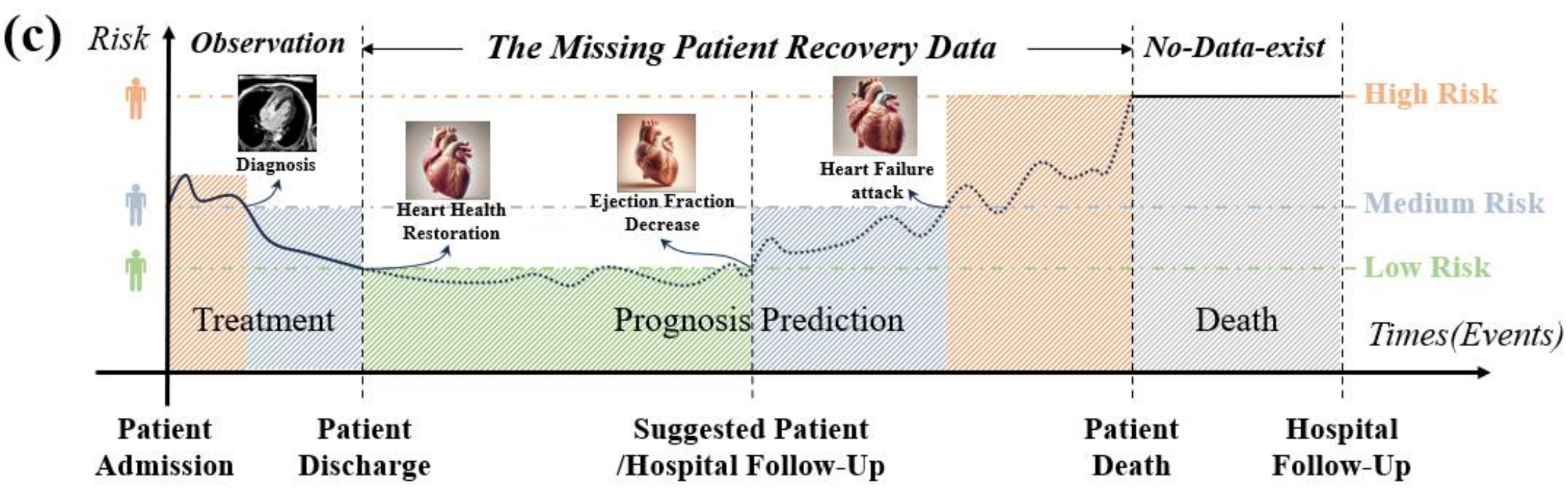

**Figure 4.** Representative Prediction Results of the Proposed Framework. (a) Example of myocardial fibrosis prediction results. The original

cine CMR image, reference fibrosis label, and predicted fibrosis region show close visual agreement. (b) Representative examples of clinical outcome prediction, including death, cause of death, cardiology rehospitalization, and MACCE-related outcomes. (c) Schematic timeline of patient risk stratification and prognosis prediction. The framework dynamically integrates multimodal information to support individualized risk assessment and follow-up planning.

By integrating multimodal data, the M-PRTM evaluates patient risk at different stages and forecasts critical events. For instance, during the treatment phase, it identifies the recovery time point for HF (green low-risk region). As the patient transitions to the follow-up stage post-discharge, potential risks such as declining ejection fraction (blue medium-risk region) begin to emerge. The M-PRTM further predicts high-risk events, such as HF deterioration (orange high-risk region), and intelligently infers risk trends during data-absent follow-up windows based on historical data. This mechanism, depicted in Fig. 4(c), demonstrates the capability of M-PRTM to support personalized follow-up strategies. It provides tailored recommendations for re-evaluation and follow-up visits, reducing unnecessary medical burdens for low-risk patients while optimizing resources and offering scientific guidance for precise interventions. We further evaluate the model on the independent test set using sensitivity, specificity, and AUPRC. Detailed metrics are presented in Table 5.

### 3.2 M-PRTM's Evaluation and Analysis

In this section, we evaluate the performance of the proposed M-PRTM under different modality combinations and analyze the impact of attention allocation strategies. When integrating textual, cine CMR, and numerical modalities, the M-PRTM achieves a maximum accuracy of 96.5%, indicating that combining multimodal data significantly enhances predictive performance. When using only two modalities, the combination of textual and cine CMR modalities achieves an accuracy of 94.0%, outperforming the combinations of textual and numerical modalities (92.5%) and cine CMR and numerical modalities (86.5%), as shown in Fig. 5(a). This highlights the critical role of the textual modality, as it contains prescription information from physicians, which is pivotal for subsequent predictions. We further investigate the impact of different attention allocation strategies on model performance, as depicted in Fig. 5(b).

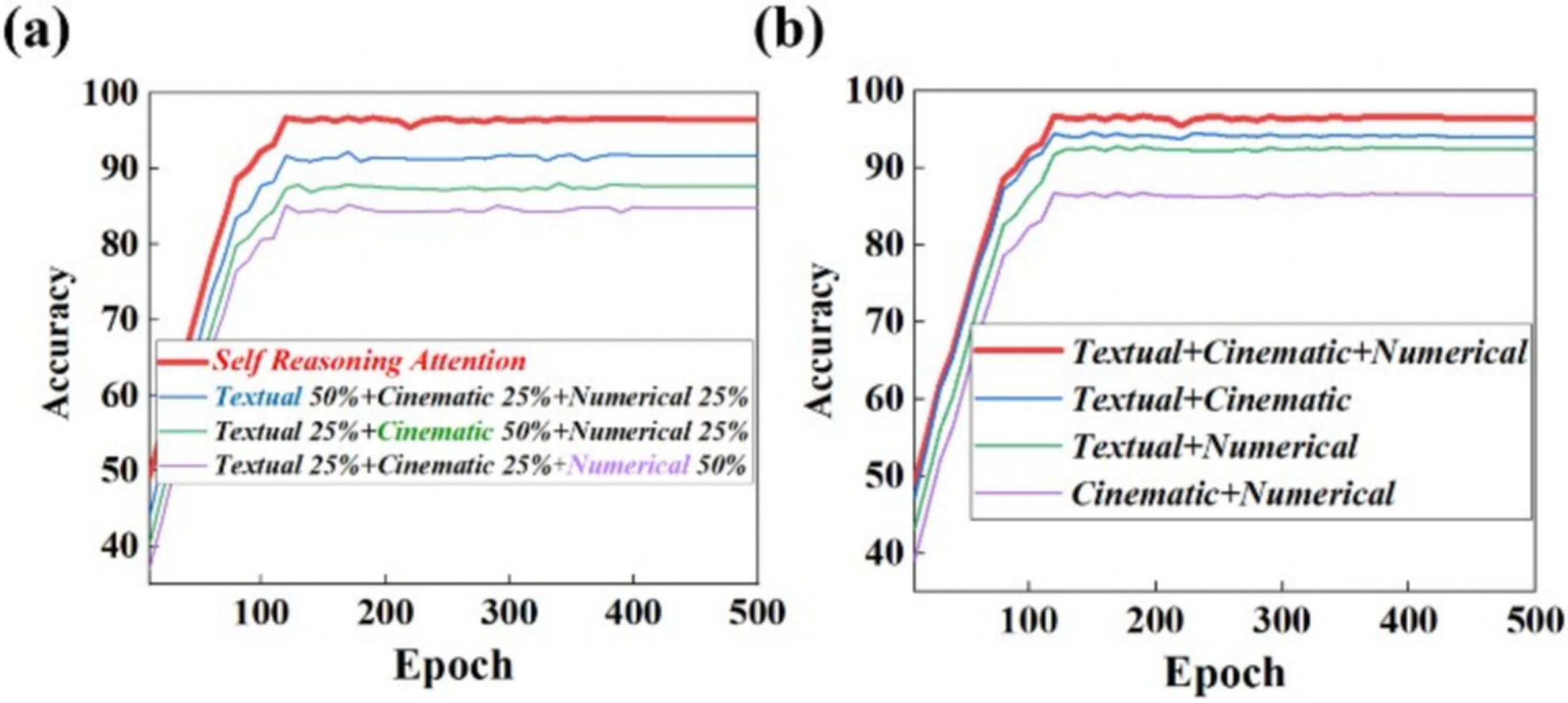

**Figure 5.** Model Performance with Attention Allocation and Modal Combinations. (a) Accuracy comparison for different modal combinations. The combination of textual, cine CMR, and numerical data achieves the highest accuracy, followed by textual and cine CMR. (b) Accuracy comparison under different attention allocation strategies. Self-Reasoning Attention achieves the best performance, outperforming fixed weight allocations.

The Self-Reasoning Attention mechanism delivers the highest accuracy of 96.5%, demonstrating its effectiveness in balancing the contributions of each modality to the predictions. Under fixed weight allocation strategies, the M-PRTM attains 91.7% accuracy with 50% weight on textual, 25% on cine CMR, and 25% on numerical modalities. This significantly outperforms allocations such as 25% textual, 50% cine CMR, and 25% numerical (87.7%), or 50% numerical alone (84.8%). It is important to clarify that the fixed weight allocation (Textual 50%, Cine CMR 25%, Numerical 25%) is selected as a static baseline to illustrate the performance limitations of manual weight assignment. This comparison serves to highlight the trend of model improvement and the superiority of the self-reasoning Attention mechanism, which dynamically optimizes feature contribution rather than relying on arbitrary, pre-defined ratios. These results demonstrate that adaptive attention allocation strategies optimize the synergy among modalities and fully leverage the strengths of different data types, leading to more consistent and generalizable performance across varied patient subgroups. The analysis indicates that the textual data plays a dominant role in predictions, as it contains critical information such as phased prescription adjustments. In contrast, numerical data such as personal and basic metrics tends to diminish the importance of features like medical history. Consequently, its contribution to the M-PRTM's attention allocation is relatively lower. Ultimately, our Self-Reasoning Attention mechanism allocates approximately 55% to textual data, 30% to cine CMR, and 15% to numerical data, as illustrated in Table 4.

**Table 4.** Performance of Modal Combinations and Attention Allocation Strategies

| Modal combination | Accuracy% | Attention allocation | Accuracy% |
|---|---|---|---|
| Textual + Cine CMR + Numerical | 96.5 | Self-attention | 96.5 |
| Textual + Cine CMR | 94.0 | Text 50% + Cine 25% +Num 25% | 91.7 |
| Textual + Numerical | 92.5 | Text 25% + Cine 50% +Num 25% | 87.7 |
| Cine CMR + Numerical | 86.5 | Text 25% + Cine 25% +Num 50% | 84.8 |

These findings further highlight the significant variation in contributions among modalities, underscoring the necessity of dynamic allocation strategies to optimize balance and maximize the model performance. However, it is important to note that these attention weights reflect the model's internal focus rather than causal feature importance (Pavlov *et al* 2024), and should be interpreted strictly as modality contribution within the inference context. Here, we analyse the attention weights for two representative patients. For Patient A (standard progression), the model assigns weights close to the global average (Text: ~55%, Cine CMR: ~30%). However, for Patient B, who presents with significant myocardial fibrosis but stable clinical metrics, the Self-Reasoning Attention mechanism intelligently increases the cine CMR weight to 82%. This adjustment allows the model to correctly predict a high-risk MACCE event by prioritizing the critical imaging features, demonstrating that the model adapts to individual patient conditions rather than relying on static modalities. To empirically validate this reliance, we conduct a Permutation Importance analysis, where shuffling textual features cause the most significant performance drop (Accuracy -12%), confirming the critical role of the textual modality in the model's decision-making. To further validate the superior performance of M-PRTM, we conduct a comparative analysis against two baseline models: a Simple Multimodal Network (which uses direct feature concatenation without attention mechanisms) and a Logistic Regression model (constructed in the style of the MAGGIC risk score using standard clinical variables). The results are summarized in Table 5. The 95% confidence intervals are estimated using bootstrap resampling on the independent test set. We further evaluate the model using comprehensive metrics on the independent test set.

**Table 5.** Performance Comparison with Baselines

| Method | Accuracy | Sensitivity ↑ | Specificity ↑ | AUPRC ↑ |
|---|---|---|---|---|
| M-PRTM (Ours) | 0.965<br>(95% CI: [0.951, 0.977]) | 0.783<br>(95% CI: [0.741, 0.821]) | 0.858<br>(95% CI: [0.828, 0.885]) | 0.814<br>(95% CI: [0.782, 0.843]) |
| Simple Multimodal Network<br>(Suzuki et al 2016) | 0.748<br>(95% CI: [0.718, 0.779]) | 0.671<br>(95% CI: [0.627, 0.716]) | 0.769<br>(95% CI: [0.748, 0.805]) | 0.723<br>(95% CI: [0.687, 0.756]) |
| Logistic Regression (MAGGIC-style)<br>(Mafort Rohen *et al* 2022) | 0.872<br>(95% CI: [0.844, 0.896]) | 0.734<br>(95% CI: [0.692, 0.772]) | 0.817<br>(95% CI: [0.785, 0.845]) | 0.793<br>(95% CI: [0.761, 0.822]) |

Note: 95% confidence intervals are estimated using bootstrap resampling on the independent test set.

As observed in Table 5, M-PRTM outperforms the Simple Multimodal Network significantly across all metrics, confirming the effectiveness of the Self-Reasoning Attention mechanism over simple concatenation. Furthermore, compared to the clinical standard Logistic Regression model, M-PRTM achieves higher Accuracy (0.965 vs. 0.872) and Sensitivity (0.783 vs. 0.734). While Specificity remains comparable due to the prevalence of negative samples in the dataset, the overall performance advantage of M-PRTM verifies the efficacy of our proposed framework.

### 3.3 Time-to-event Regression Performance

To further assess the auxiliary time-to-event prediction branch, we evaluate its regression performance on patients with observed MACCE events and valid event-time annotations. The model achieves a mean absolute error (MAE) of **14.3** days (95% CI: [11.2, 17.8]), a root mean squared error (RMSE) of **22.6** days (95% CI: [18.5, 27.2]), and an $R^2$ of **0.64** (95% CI: [0.57, 0.71]). Specifically, the MAE indicates an average prediction error of approximately two weeks, providing a rough clinical reference for follow-up scheduling. The RMSE, which is slightly larger than the MAE, reflects occasional large errors, which is acceptable for a high-risk cohort. The $R^2$ suggests that the model explains 64% of the variance in event timing, indicating moderate predictive ability with room for improvement. This finding is consistent with the auxiliary role of the regression branch rather than positioning it as a substitute for formal survival analysis. These results indicate that the regression branch provides a preliminary estimate of event timing for high-risk patients. From a clinical perspective, this component may serve as a supplementary reference for follow-up planning and risk communication, rather than as a replacement for formal survival analysis.

## 4. Conclusion

### 4.1 Methodological Innovations Clinical Impact

In this study, we introduce the M-PRTM, a Composable Strategy Framework for HF prognosis that leverages multimodal clinical intelligence. Our model integrates three primary data modalities: numerical indicators, textual prescriptions, and cine CMR. Each modality is processed using specialized architectures, with the DAE-Former for cine CMR, a fully connected network for numerical data, and a BERT-based model for textual data. These representations are fused through an adaptive attention mechanism (Sukhbaatar *et al* 2019, Liu *et al* 2021) that dynamically prioritizes critical features, such as drug prescriptions and vital signs. The model achieves an accuracy of 96.5% in HF prognosis prediction, highlighting the benefits of multimodal integration for clinical decision support.

Our findings reveal several key insights into the role of multimodal data in HF assessment. The fusion of textual, numerical, and imaging data provides a more comprehensive evaluation of patient status, compensating for the limitations of single-modality approaches. Notably, textual prescription data proves essential in predicting clinical outcomes, particularly in forecasting rehospitalization risk and determining optimal follow-up strategies. The framework captures treatment trends that significantly influence patient recovery by analyzing longitudinal drug usage patterns. Additionally, the DAE-Former model processes CMR data to enhance myocardial fibrosis detection, which serves as a key prognostic factor for declining heart function. Multimodal data integration improves predictive accuracy and supports personalized treatment, highlighting the value of cross-domain feature fusion in medical diagnostics.

### 4.2 Clinical Impact

The integration of multimodal data holds great promise for optimizing clinical management, particularly in complex cases like heart failure. The model's dynamic prioritization of critical features, such as drug prescriptions and vital signs, supports personalized treatment strategies and enables early intervention. By accurately predicting key clinical outcomes such as rehospitalization risk and myocardial fibrosis progression, the M-PRTM can facilitate better resource allocation, reducing unnecessary follow-up visits for low-risk patients while guiding timely interventions for high-risk individuals.

Specifically, regarding the deployment scenario, M-PRTM is designed to be utilized at the critical transition point of patient discharge. The model leverages comprehensive data available from the index hospitalization—including medical history, laboratory reports, baseline cine CMR, and the finalized discharge medication regimen—to predict whether a key outcome (MACCE) will occur within the subsequent 24 months, along with its specific type and severity. This predictive capability offers substantial value for prognosis assessment. When a heart failure patient is discharged, the hospital can leverage our network to generate a 24-month risk trajectory. The detection of a high-risk profile allows the clinical team to quantify the likelihood of deterioration and implement preemptive interventions (e.g., closer follow-up intervals or optimized titration) rather than waiting for symptoms to worsen. This early anticipation of disease progression is crucial for reducing mortality risks associated with delayed treatment. Furthermore, in a prospective setting, physicians could potentially utilize the model's risk feedback to adjust treatment plans *during* the hospitalization phase to achieve optimal therapeutic outcomes, which represents a key direction for our future work. In particular, the model's ability to predict the course of myocardial fibrosis, a key prognostic factor for heart failure, positions it as a valuable tool in the clinical evaluation of heart failure patients.

### 4.3 Challenges and Future Directions

Despite its strong performance, certain limitations in the dataset may affect the M-PRTM's generalizability. The dataset used in this study originates from a single medical institution, which may not fully represent the diversity of patient populations across different geographical and clinical settings. Expanding the dataset to include multi-center cohorts with broader demographic and pathological variations is essential to improve the robustness. Additionally, challenges related to data quality, such as missing values, inconsistent clinical records, and variations in imaging protocols, should be addressed. Future work should focus on implementing advanced preprocessing techniques and data augmentation strategies to mitigate these issues. Furthermore, refining the M-PRTM's ability to handle incomplete data will be crucial for its real-world applicability in heterogeneous clinical environments.

While our framework demonstrates high predictive performance, its interpretability remains a challenge, particularly given the complexity of deep learning-based clinical models. The "black box" nature of neural networks poses obstacles for clinical adoption, as medical professionals require transparency in decision-making. Although our attention mechanism dynamically adjusts modality importance, further efforts are necessary to enhance the explainability. Future work should explore post-hoc interpretability methods and integrate expert-driven validation, such as comparisons with cardiologist assessments, may further enhance trust and acceptance in real-world clinical settings.

Looking ahead, the M-PRTM framework shows great potential beyond HF prognosis. Its integration of cine CMR, textual, and numerical data allows adaptation to other medical fields, such as oncology, diabetes management, and neurodegenerative disease monitoring. In resource-limited settings, where advanced imaging may be less accessible, the model's reliance on textual and numerical data still enables highly accurate predictions, expanding its applicability. As multimodal data collection and AI methodologies continue to evolve, frameworks like M-PRTM are expected to transform clinical decision support, driving the future of precision medicine and personalized healthcare.

### Acknowledgements

This work was supported by the National Natural Science Foundation of China under Grant 12404365. The authors thank Professor Xiaofei Li for insightful discussions.

### Data availability

Our source code is available at: https://github.com/AlexSun111111/Multimodal-Post-Recovery-Tracking-Model.

### Conflict of interest

The authors declare that they have no known competing financial interests or personal relationships that could have appeared to influence the work reported in this paper.

### Ethics approval

Ethical approval was granted by the Ethics Committee of Nanjing Drum Tower Hospital (Approval No. 2025-0094-015).

### Consent to participate

Not applicable.

### Consent to publication

All authors have checked the manuscript and have agreed to the submission.